\documentclass{article}
\usepackage{ijcai26}

\usepackage{times}
\usepackage{soul}
\usepackage{url}
\usepackage[hidelinks]{hyperref}
\usepackage[utf8]{inputenc}
\usepackage[small]{caption}
\usepackage{graphicx}
\usepackage{amsmath}
\usepackage{amsthm}
\usepackage{booktabs}
\usepackage{algorithm}
\usepackage{algorithmic}
\usepackage[switch]{lineno}
\usepackage{multirow}
\usepackage{subcaption}

\newcommand{\modelname}{HRSino}

\title{Training-Free Inference for High-Resolution Sinogram Completion}

\author{
Jiaze E$^1$
\And
Srutarshi Banerjee$^2$
\And
Tekin Bicer$^{2,3}$
\And
Guannan Wang$^1$
\And
Yanfu Zhang$^1$
\and
Bin Ren$^1$
\\
\affiliations
$^1$William \& Mary\\
$^2$Argonne National Laboratory\\
$^3$University of Chicago\
\emails
je@wm.edu,
\{sruban, tbicer\}@anl.gov,
\{gwang01. yzhang105, bren\}@wm.edu
}
\begin{document}

\maketitle

\begin{abstract}
    High-resolution sinogram completion is critical for computed tomography reconstruction, as missing projections can introduce severe artifacts. While diffusion models provide strong generative priors for this task, their inference cost grows prohibitively with resolution. We propose ~\modelname, a training-free and efficient diffusion inference approach for high-resolution sinogram completion. By explicitly accounting for spatial heterogeneity in signal characteristics, such as spectral sparsity and local complexity, ~\modelname~allocates inference effort adaptively across spatial regions and resolutions, rather than applying uniform high-resolution diffusion steps. This enables global consistency to be captured at coarse scales while refining local details only where necessary. Experimental results show that~\modelname~reduces peak memory usage by up to 30.81\% and inference time by up to 17.58\% compared to the state-of-the-art framework, and maintains completion accuracy across datasets and resolutions.
\end{abstract}

\section{Introduction}

A sinogram is a 2D projection-domain representation of computed tomography (CT) data, where each row corresponds to an acquisition angle and each column to a detector position. In practical CT reconstruction pipelines, projection data are often incomplete due to limited-angle acquisition, reduced sampling, or experimental constraints, and missing sinogram measurements can introduce severe artifacts that are amplified during reconstruction \cite{kalender2011computed}. Sinogram completion therefore plays a critical role in restoring projection data prior to image reconstruction.

High-resolution sinograms, often reaching 2048$\times$2048 or beyond, arise in synchrotron and nano-scale CT systems, where dense angular sampling and fine detector resolution are required to preserve quantitative accuracy. In these settings, high resolution is required to preserve measurement fidelity and ensure stable reconstruction, and aggressive resolution reduction can fundamentally alter the inverse problem and compromise downstream analysis \cite{slaney1988principles}.

\begin{figure}
  \centering
  \includegraphics[width=0.92\linewidth]{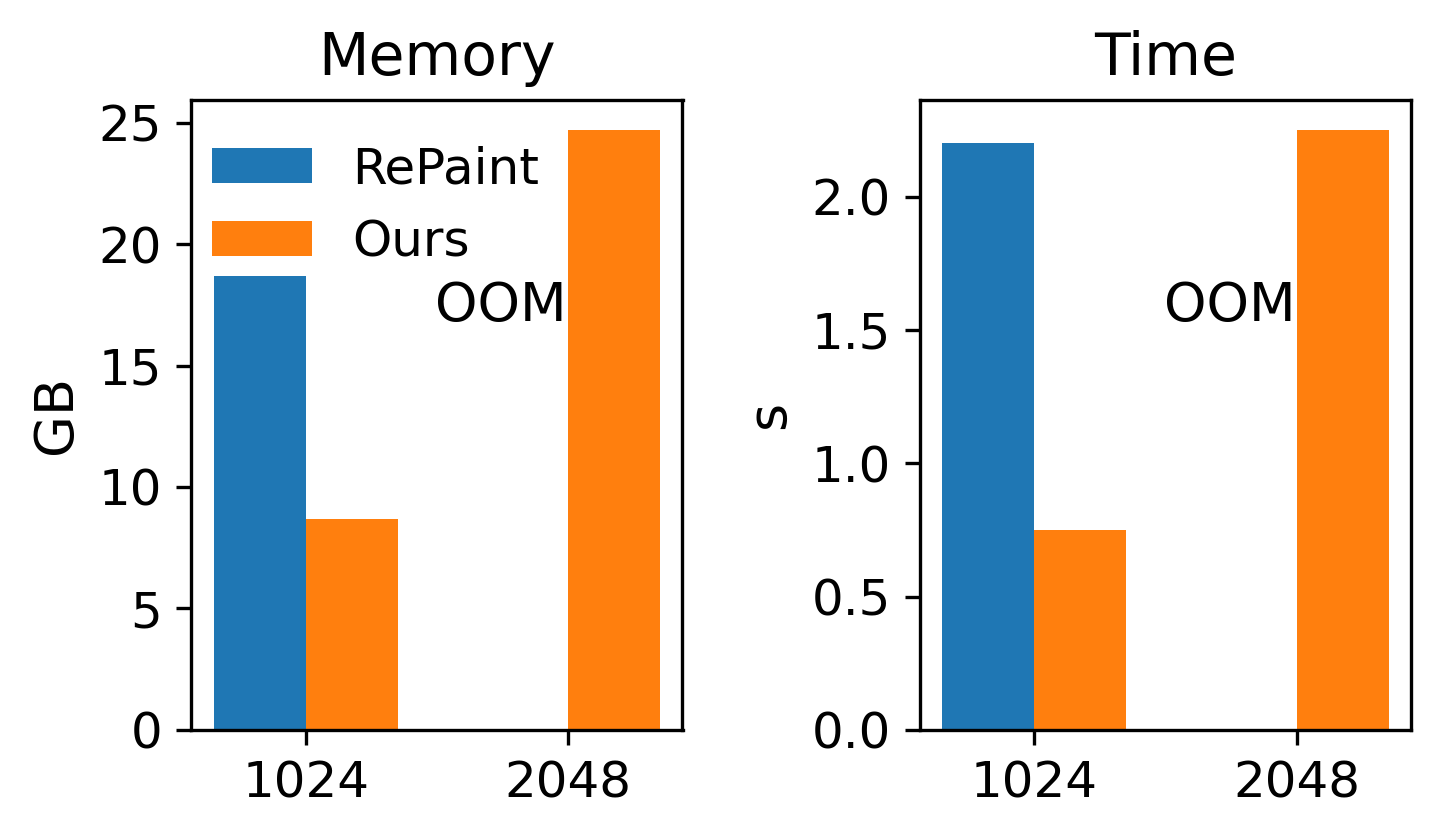}
  \caption{Peak GPU memory usage (left) and inference time (right) for sinogram completion at 1024$\times$1024 and 2048$\times$2048 resolutions. Unoptimized diffusion inference encounters out-of-memory (OOM) at high resolution, while our approach remains feasible and substantially reduces memory consumption and runtime.}
  \label{fig:intro}
\end{figure}

Under the high-resolution setting described above, inference efficiency becomes a primary bottleneck for sinogram completion. Diffusion models have demonstrated strong performance in completion due to their ability to progressively refine structure through iterative denoising \cite{sohl2015deep,ho2020denoising,saharia2022palette,lugmayr2022repaint}. However, this inference procedure requires repeatedly applying a large UNet over the full spatial domain across many denoising steps, leading to substantial memory and computational overhead as resolution increases. In practical high-resolution sinogram settings, this overhead translates into tens of gigabytes of GPU memory consumption and seconds of inference time, even when processing a single input at a time. Such costs quickly exceed the capacity of commonly available hardware, making high-resolution sinogram deployment challenging. As illustrated in Fig~\ref{fig:intro}, unoptimized diffusion inference exhibits prohibitive memory consumption and runtime at 2048$\times$2048 resolution, resulting in out-of-memory (OOM) failures.

A number of approaches have been proposed to improve diffusion efficiency. Step reduction and sampling acceleration methods aim to shorten the denoising trajectory by modifying the global sampling schedule \cite{lu2022dpm,salimans2022progressive,meng2023distillation}. Other efforts focus on architectural or implementation-level optimizations to reduce per-step computation or memory usage \cite{li2023snapfusion,zhang2024effortless,zhu2024dip}. While effective in reducing overall inference cost, these methods typically apply computation uniformly across the entire spatial domain and maintain a fixed denoising depth for all regions. As a result, computational effort is often wasted on regions that require little refinement, leaving substantial redundancy in high-resolution diffusion inference under strict memory constraints. This limitation becomes particularly pronounced in high-resolution sinogram completion, where large spatial regions exhibit low signal variation and do not require uniform refinement across all diffusion steps.

Motivated by this perspective, we propose~\modelname, a training-free efficient high-resolution sinogram completion framework that reorganizes diffusion inference along both spatial and temporal dimensions by selectively allocating computation across regions and adapting denoising depth to local complexity. Rather than modifying model architecture or retraining diffusion networks,~\modelname~operates entirely at inference time by reallocating computation to where it is most needed. To enable patch-wise diffusion inference without full-frame activation, we employ a resolution-guided inference scheme that extracts coarse global structure at a reduced resolution and uses it to guide localized refinement at higher resolutions. By avoiding denoising on the full-resolution grid from the outset, this strategy substantially reduces peak memory usage while preserving global consistency. Unlike prior approaches that introduce global embeddings, conditioning tokens, or architectural modifications to maintain context \cite{avrahami2022blended,zhang2023towards,tumanyan2023plug,zhang2023adding}, this mechanism operates entirely at inference time and requires no retraining.

In practice, a significant portion of redundancy in high-resolution diffusion inference comes from large variations in local signal complexity across the spatial domain. Large portions of the input often exhibit smooth variation and limited high-frequency content \cite{slaney1988principles}, yet are subjected to the same inference cost as structurally rich regions. To address this imbalance,~\modelname~introduces a frequency-aware patch-skipping mechanism that identifies low-energy patches using a simple FFT-based score. Patches with negligible spectral content are bypassed during diffusion inference and approximated using compact representations, thereby eliminating redundant computation without compromising completion fidelity.

For the remaining patches, structural complexity can vary substantially, ranging from flat gradients to fine edges and high-frequency details. Applying a fixed number of denoising steps to all patches therefore leads to either over-computation in simple regions or insufficient refinement in complex ones. To better align inference cost with local content richness,~\modelname~employs a structure-adaptive denoising scheduler that assigns each patch a customized number of denoising steps. This allocation is determined by a joint complexity score combining Shannon entropy \cite{shannon1948mathematical} and frequency energy, capturing both spatial variability and spectral content. Together, frequency-aware patch skipping and adaptive step allocation enable diffusion inference to concentrate computational effort on structurally informative regions, while substantially reducing redundant computation under strict constraints.

Together, these designs preserve completion fidelity by first capturing global structure that maintains long-range consistency across resolution levels. Patch skipping then removes only regions with negligible signal, ensuring that no structurally relevant information is lost, while adaptive denoising assigns each patch a sufficient number of steps based on its complexity, ensuring adequate refinement in detail-rich areas. As a result,~\modelname~improves efficiency (both in peak memory usage and inference speed) without compromising completion quality. Overall, our contributions are as follows:
\begin{itemize}
    \item We present~\modelname, a novel inference-time diffusion framework for high-resolution sinogram completion under strict constraints. The framework reorganizes diffusion inference to enable efficient high-resolution completion without retraining or architectural modifications.
    \item We introduce a pair of inference-time mechanisms: frequency-aware patch skipping and structure-adaptive denoising, which enable spatially adaptive allocation of diffusion computation. By selectively skipping low-information regions and assigning different denoising depths based on local complexity, these mechanisms address the redundancy inherent in uniform full-resolution diffusion inference.
    \item \modelname~enables 2048$\times$2048 sinogram completion on an A100 GPU, reducing peak memory usage by up to 30.81\% and inference time by up to 17.58\% without degrading completion quality across multiple datasets, input resolutions, mask types, and mask ratios.
\end{itemize}

\section{Related Work}

\noindent{\bf Deep Learning Based Completion Methods.} Early approaches employ CNN-, GAN-, or transformer-based architectures. Representative examples include ICT \cite{wan2021high} which leverages transformer attention to model long-range structures and produces pluralistic high-fidelity completion results. Recently, diffusion-based models have achieved strong performance on image completion due to their generative capacity and ability to model uncertainty. Methods such as RePaint \cite{lugmayr2022repaint}, Palette \cite{saharia2022palette}, Blended Diffusion \cite{avrahami2022blended}, and CoPaint \cite{zhang2023towards} enhance global coherence on images through iterative sampling or semantic priors. Sinogram completion has also been studied in industrial and medical CT settings using CNNs \cite{lee2018deep,jin2017deep} and transformer-based such as SinoTx \cite{jiaze2025sinotx} for sparse-view recovery or artifact reduction, but these works generally ignore memory constraints and inference time optimizations. End-to-end CT reconstruction methods \cite{guo2025advancing} directly map incomplete sinograms to reconstructed images. While effective for reducing image-space artifacts, they do not provide sinogram-domain control or modular compatibility with existing CT reconstruction workflows.

\noindent{\bf Inference Acceleration in Diffusion Models.}
Improving the efficiency of diffusion inference has become a growing area of focus. Sampling-based methods reduce denoising steps through distillation \cite{salimans2022progressive}, or step-aware training \cite{xiao2024fastcomposer}. Architectural methods employ memory-aware attention, such as in HiDiffusion \cite{zhang2024hidiffusion} and DiffIR \cite{xia2023diffir}. Model pruning and lightweight design strategies have also been proposed, as seen in SnapFusion \cite{li2023snapfusion}, EcoDiff \cite{zhang2024effortless}, and DiP-GO \cite{zhu2024dip}. Patch-wise generation and conditional masking \cite{avrahami2023blended} have also been explored to control computation scope. However, most existing methods require retraining, loss rebalancing, or internal modifications to the model. Many of these approaches are orthogonal to our work and can be integrated in principle, but they do not account for and apply to the structural characteristics, such as directional sparsity and local interpretability. Applying them effectively would require additional adaptation. 

\noindent{\bf Memory and Runtime Efficient Deep Learning.}
Beyond diffusion, memory- and latency-aware inference strategies have been widely studied in deep learning. Techniques include recomputation and memory reuse \cite{chen2016training,jain2020checkmate}, and runtime-adaptive depth scaling \cite{xia2021fully}. MEST \cite{yuan2021mest} also explore sparsity-aware training or deployment under tight memory budgets. These techniques address memory bottlenecks in standard feedforward networks and our work focuses instead on iterative generative models with different constraints.

\section{\modelname: High-Resolution Sinogram Completion}

\begin{figure*}
  \centering
  \includegraphics[scale=0.25]{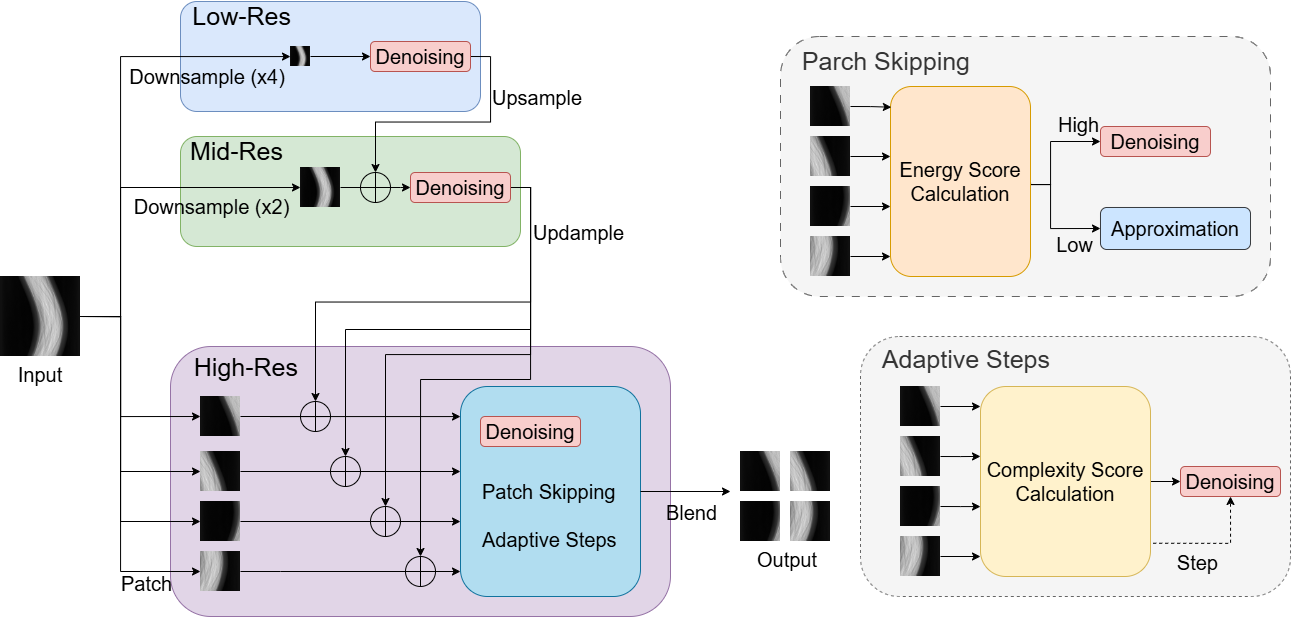}
  \caption{Overview of~\modelname. The left illustrates the three-stage resolution-guided pipeline: low resolution followed by mid and high resolutions in a progressive refinement scheme, with the final stage performed patch-wise for detail recovery. At each stage, the upsampled output from the previous resolution is fused with the current input before denoising. The right zooms into the high-resolution stage, where two modules are applied: frequency-aware patch skipping (bypassing low-information patches) and structure-adaptive step allocation (assigning variable denoising steps by patch complexity).}
  \label{fig:overview}
\end{figure*}

As shown in Fig~\ref{fig:overview},~\modelname~builds upon RePaint \cite{lugmayr2022repaint} as its default backbone and restructures the inference process into a three-stage resolution-guided pipeline. The sinogram is first denoised at low resolution to establish global structure, then refined at mid resolution, and finally completed at full resolution through patch-wise inference. At each stage, the upsampled output from the previous resolution is fused with the current input before denoising, ensuring hierarchical guidance across scales. This hierarchical design avoids full-frame activation and substantially reduces memory usage while preserving long-range consistency. To further improve efficiency under the structural characteristics of sinograms,~\modelname~introduces two inference-time modules: (1) frequency-aware patch skipping, which exploits the spectral sparsity of background regions to bypass redundant computation, and (2) structure-adaptive step allocation, which leverages local structural heterogeneity to adjust denoising depth per patch.

\subsection{Resolution-Guided Progressive Inference}

To efficiently complete high-resolution sinograms while avoiding memory overflow,~\modelname~adopts a three-stage progressive inference pipeline operating over low, mid, and high resolutions. Let $x_r$ denote the input at resolution $r \in \{low, mid, high\}$. Each resolution level $x_r$ is obtained by downsampling the original sinogram using a fixed ratio: low and mid-resolution inputs correspond to $0.25\times$ and $0.5\times$ the original resolution, respectively. At the first stage, full DDIM inference is performed on $x_{low}$ to generate a coarse prior $\hat{x}_{low}$ to capture global structural cues at minimal memory cost.

In the second stage, we refine the geometry at mid-resolution. $x_{mid}$ is fused with the upsampled output from the previous stage using a resolution-aware weighted sum:
\begin{equation}
    \tilde{x}_{mid} = \frac{1}{2} \cdot (Up (\hat{x}_{low}) + x_{mid}),
\end{equation}
where $Up(\cdot)$ denotes nearest-neighbor upsampling. This fusion helps the model retain mid-level structure while reinforcing global context. The resulting $\tilde{x}_{mid}$ is then denoised via DDIM to produce $\hat{x}_{mid}$.

At the final stage, we operate on the original full-resolution input. We divide $x_{high}$ into patches $\{x_{high}^i\}$. Each patch is of fixed size, ensuring consistent inference cost per patch across resolutions. For each patch, we retrieve the aligned region from $\hat{x}_{mid}$, upsample it, and fuse it with the local patch:
\begin{equation}
    \tilde{x}_{high}^i = \frac{1}{2} \cdot (Up (\hat{x}_{mid}^i) + \cdot x_{high}^i).
\end{equation}
Each $\tilde{x}_{high}^i$ is then denoised via DDIM. Sec~\ref{sec:patch-skipping} and Sec~\ref{sec:adaptive-denoising} introduce more details about the final stage. This patch-wise processing avoids full-frame memory load while preserving long-range consistency through hierarchical conditioning. Each patch is processed sequentially during inference to ensure bounded memory usage.

\subsection{Frequency-Aware Patch Skipping}
\label{sec:patch-skipping}

High-resolution sinograms often contain broad low-frequency backgrounds with narrow high-frequency details. To reduce computation in low-complexity regions,~\modelname~introduces frequency-aware patch skipping based on localized spectral content.

\subsubsection{Frequency-Aware Patch Pruning}

To reduce unnecessary computation in low-information regions of high-resolution sinograms, we introduce a mask-aware frequency-based patch skipping strategy. Let $P$ denote a high-resolution sinogram patch, and let $\mathcal{F}(P)$ represent its real-valued 2D Fourier transform. We first compute the high-frequency energy ratio $\gamma (P)$ as:
\begin{equation}
    \gamma (P) = \frac{\sum_{(u,v) \in \Omega_{high}} |\mathcal{F}(P)_{u,v}|^2}{\sum_{(u,v)} |\mathcal{F}(P)_{u,v}|^2},
\end{equation}
where $\Omega_{high}$ is a predefined high-pass band, typically the outer third of the frequency spectrum. In addition, we measure the mask ratio $r$. To incorporate mask awareness, we define the adjusted score as:
\begin{equation}
    \gamma^{\prime} (P) = (1-r(P)) \cdot \gamma (P) + \tau \cdot r(P),
\end{equation}
where $\tau$ is a small constant that safeguards against skipping heavily masked patches. A patch is considered redundant and skipped if its adjusted score $\gamma^{\prime} (P)$ falls below a threshold $\tau$. In this way, structurally smooth or empty regions can still be skipped, while patches with large missing areas are preserved for reliable completion.

Rather than running full DDIM inference for spectrally sparse patches, we replace their outputs with a fixed approximation. It is obtained by passing a synthetic input patch—filled with low-amplitude Gaussian noise with mean 0, and standard deviation 0.01—through the original RePaint and DDIM steps. This design simulates typical background regions. It avoids repeated computation in trivial regions and ensures consistency with standard outputs.

\subsubsection{Cosine-Based Patch Blending}

To avoid visual artifacts at patch boundaries, particularly when adjacent regions are inferred with different mechanisms, we apply a smooth blending operation between neighboring patches. This is especially important in high-resolution sinograms where directional continuity must be preserved.

Let $P_1$ and $P_2$ denote two horizontally adjacent reconstructed patches, and $p \in [0,L]$ be local coordinate across boundary region of width $L$ pixels. Define a cosine-based spatial weight function:
\begin{equation}
    \alpha (p) = \frac{1}{2} \left(1-cos(\frac{\pi p}{L})\right),
\end{equation}
which increases smoothly from 0 to 1 across the boundary. The blended pixel value is computed as:
\begin{equation}
    P_{blend} (x,y) = \alpha (p) \cdot P_1 (x,y) + (1-\alpha (p)) \cdot P_2 (x,y).
\end{equation}
This produces a smooth interpolation where the transition is soft and visually imperceptible, reducing abrupt intensity jumps at patch edges.

We apply this blending only when necessary. Specifically, we compute the average gradient magnitude in the overlapping region using a standard Sobel filter applied to the corresponding mid-resolution sinogram. If the gradient exceeds a predefined threshold $\eta$, blending is enabled; otherwise, a simple hard stitch is used to save computation in flat areas. This conditional mechanism ensures efficiency without compromising visual consistency in structurally complex regions.

\subsection{Structure-Adaptive Denoising}
\label{sec:adaptive-denoising}

To further reduce computation, we allocate a different number of denoising steps to each patch based on its structural complexity. Unlike static diffusion sampling where every region receives the same number of DDIM steps, we adapt the inference depth dynamically to match the content difficulty.

\subsubsection{Complexity Score Computation}

Let $P$ denote a patch in the high-resolution sinogram. We define a complexity score $\kappa (P)$ that combines two frequency-domain measures: Shannon entropy and high-frequency energy. The Shannon entropy $\mathcal{H}(P)$ is computed over the pixel intensity histogram:
\begin{equation}
    \mathcal{H} (P_i) = - \sum_j p_j \log p_j,
\end{equation}
where $p_j$ is the normalized count of pixel intensity in bin $i$. This captures texture randomness and distribution uniformity.

Also, we compute the spectral energy using the L1-norm of the 2D FFT of the patch. The overall complexity score is defined as:
\begin{equation}
    \kappa_i = \mathcal{H} (P_i) + \log (1 + ||\mathcal{F} (P_i)||_1),
\end{equation}
where $\mathcal{F} (P_i)$ is the 2D real-valued FFT of the patch and $|| \cdot ||_1$ denotes the L1-norm of the spectrum.

Entropy captures randomness and texture variation in the spatial domain, while the FFT energy captures signal richness and directional complexity. Their combination provides a robust, domain-agnostic proxy for structural difficulty.

\subsubsection{Patch-Wise Step Mapping}

\begin{table}[t]
    \centering
    {\small
    \begin{tabular}{lccccc}
        \toprule
        \multicolumn{1}{l}{\multirow{2}{*}{Method}} & \multicolumn{2}{c}{2048$\times$2048} & \multicolumn{2}{c}{1024$\times$1024} \\
        & Mem & Inf time & Mem & Inf time \\
        \midrule
        RePaint & \multicolumn{2}{c}{OOM} & 18.7 & 2.20 \\
        MCG & \multicolumn{2}{c}{OOM} & 23.2 & 2.81 \\
        DiffIR & \multicolumn{2}{c}{OOM} & 22.0 & 2.54 \\
        TD-Paint & \multicolumn{2}{c}{OOM} & 19.3 & 1.48 \\
        HiDiffusion & 35.7 & 2.73 & 12.5 & 0.92 \\
        HRSino (ours) & \textbf{24.7} & \textbf{2.25} & \textbf{8.7} & \textbf{0.75} \\
        \midrule
        SinoTx & \multicolumn{2}{c}{OOM} & 19.5 & 0.38 \\
        ICT & \multicolumn{2}{c}{OOM} & 17.0 & 0.20 \\
        \midrule
        Spline (CPU) & 0.4 & 0.22 & 0.4 & 0.07 \\
        Bilinear (CPU) & 0.3 & 0.12 & 0.3 & 0.04 \\
        \bottomrule
    \end{tabular}}
    \caption{Peak GPU memory (GB) and inference time (s) of different methods on {\tt TomoBank} with random masks (ratio = 0.8) at resolutions 2048$\times$2048 and 1024$\times$1024. “OOM” indicates out-of-memory. Results on {\tt LoDoPaB} exhibit nearly identical patterns in both memory and runtime and are omitted here for brevity.}
    \label{tab:memory-runtime}
\end{table}

To convert $\kappa_i$ into a patch-specific number of denoising steps, we apply a sigmoid-based scaling function:
\begin{equation}
    S_i = \lfloor S_{min} + (S_{max} - S_{min}) \cdot \sigma  (\kappa_i - \mu)) \rfloor
\end{equation}
where $\mu$ is the mean complexity across all patches in the sinogram; $\sigma (\cdot)$ is the standard sigmoid function; $S_{min}$, $S_{max}$ define the step range; $\lfloor \cdot \rfloor$ denotes rounding down to the nearest integer.

This ensures a soft but data-driven allocation: patches with below-average complexity receive fewer sampling steps, while others are modeled more deeply. Using the sigmoid ensures that step variation remains smooth, differentiable, and robust to outliers. The centering around the mean further normalizes the step distribution across sinograms with varying overall complexity.

\section{Evaluation}
\label{sec:eva}

The evaluation has two main objectives: (1) demonstrating that~\modelname~significantly improves memory efficiency and inference speed while maintaining completion quality (Sec~\ref{sec:overall}); (2) conducting ablation studies to validate the effectiveness of our mechanisms, including multi-resolution progressive inference, frequency-aware patch skipping and structure-adaptive denoising (Sec~\ref{sec:ablation}).

\subsection{Experimental Setup}
\label{sec:setup}

\begin{table*}
    \centering
    {\scriptsize
    \begin{tabular}{llcccc}
        \toprule
        \multirow{2}{*}{Method} & \multirow{2}{*}{Mask} & \multicolumn{2}{c}{TomoBank} & \multicolumn{2}{c}{LoDoPaB} \\
        \cmidrule(r){3-4} \cmidrule(r){5-6} &  & Ratio = 0.6 & Ratio = 0.8 & Ratio = 0.6 & Ratio = 0.8 \\
        \midrule
        \multicolumn{6}{c}{2048$\times$2048} \\
        \midrule
        \multirow{2}{*}{HRSino} & Random & \textbf{0.930} (0.916) / \textbf{30.9} (29.9) & \textbf{0.927} (0.913) / \textbf{30.6} (29.7) & \textbf{0.940} (0.926) / \textbf{31.6} (30.4) & \textbf{0.935} (0.924) / \textbf{31.3} (30.4)\\
        & Periodic & \textbf{0.926} (0.913) / \textbf{30.6} (29.7) & \textbf{0.924} (0.910) / \textbf{30.3} (29.4) & \textbf{0.937} (0.922) / \textbf{31.3} (30.4) & \textbf{0.934} (0.924) / \textbf{31.0} (30.1)\\
        \multirow{2}{*}{HiDiffusion} & Random & 0.903 (0.889) / 29.1 (28.2) & 0.900 (0.886) / 28.8 (27.9) & 0.911 (0.899) / 29.8 (28.9) & 0.910 (0.893) / 29.5 (28.6) \\\
        & Periodic & 0.902 (0.887) / 28.9 (28.0) & 0.897 (0.883) / 28.6 (27.7) & 0.912 (0.898) / 29.6 (28.7) & 0.904 (0.892) / 29.1 (28.3) \\
        \midrule
        \multirow{2}{*}{Spline} 
        & Random & 0.702 (0.688) / 23.8 (22.9) & 0.698 (0.684) / 23.5 (22.6) & 0.714 (0.694) / 24.5 (23.6) & 0.711 (0.690) / 24.3 (23.4)\\
        & Periodic & 0.698 (0.681) / 23.7 (22.6) & 0.694 (0.680) / 23.2 (22.3) & 0.709 (0.694) / 24.4 (23.5) & 0.703 (0.690) / 23.9 (23.1)\\
        \multirow{2}{*}{Bilinear} 
        & Random & 0.695 (0.681) / 22.5 (21.7) & 0.691 (0.677) / 22.3 (21.4) & 0.703 (0.695) / 23.3 (22.6) & 0.703 (0.687) / 22.9 (22.2)\\
        & Periodic & 0.695 (0.677) / 22.1 (21.4) & 0.687 (0.673) / 21.9 (21.1) & 0.703 (0.686) / 22.8 (22.0) & 0.695 (0.681) / 22.3 (21.5)\\
        \midrule
        \multicolumn{6}{c}{1024$\times$1024} \\
        \midrule
        \multirow{2}{*}{HRSino} & Random & \textbf{0.932} (0.919) / \underline{31.2} (30.4) & \underline{0.927} (0.911) / \textbf{30.9} (30.0) & \textbf{0.944} (0.929) / \textbf{32.1} (31.3) & \textbf{0.936} (0.924) / \underline{31.1} (30.5) \\
        & Periodic & \textbf{0.929} (0.916) / \underline{30.9} (30.1) & \underline{0.924} (0.913) / \underline{30.6} (29.8) & \underline{0.937} (0.924) / \underline{31.5} (30.8) & \underline{0.935} (0.922) / \underline{30.9} (30.2) \\
        \multirow{2}{*}{RePaint} & Random & \textbf{0.932} (0.919) / \textbf{31.4} (30.6) & \textbf{0.928} (0.915) / \textbf{30.9} (30.2) & \textbf{0.944} (0.927) / \textbf{32.1} (31.4) & \textbf{0.936} (0.925) / \textbf{31.2} (30.4) \\
        & Periodic & \textbf{0.929} (0.914) / \textbf{31.1} (30.3) & \textbf{0.925} (0.912) / \textbf{30.8} (30.0) & \textbf{0.938} (0.924) / \textbf{31.8} (31.1) & \textbf{0.934} (0.924) / \textbf{31.0} (30.5) \\
        \multirow{2}{*}{HiDiffusion} & Random & 0.902 (0.888) / 29.8 (29.0) & 0.898 (0.884) / 29.4 (28.6) & 0.914 (0.900) / 30.7 (29.8) & 0.909 (0.896) / 30.3 (29.5) \\
        & Periodic & 0.899 (0.885) / 29.5 (28.7) & 0.895 (0.881) / 29.1 (28.3) & 0.910 (0.896) / 30.4 (29.6) & 0.906 (0.892) / 29.9 (29.1)\\
        \multirow{2}{*}{MCG} & Random & 0.901 (0.887) / 29.9 (29.1) & 0.896 (0.883) / 29.4 (28.7) & 0.914 (0.899) / 30.6 (29.9) & 0.908 (0.894) / 30.2 (29.5) \\
        & Periodic & 0.897 (0.884) / 29.5 (28.8) & 0.894 (0.880) / 29.1 (28.3) & 0.909 (0.895) / 30.1 (29.4) & 0.904 (0.891) / 29.8 (29.0) \\
        \multirow{2}{*}{DiffIR} & Random & 0.893 (0.881) / 29.4 (28.6) & 0.889 (0.877) / 29.0 (28.2) & 0.905 (0.892) / 30.2 (29.4) & 0.900 (0.888) / 29.8 (29.0) \\
        & Periodic & 0.890 (0.878) / 29.1 (28.3) & 0.886 (0.874) / 28.7 (27.9) & 0.901 (0.889) / 29.7 (28.9) & 0.897 (0.885) / 29.3 (28.5) \\
        \multirow{2}{*}{SinoTx} & Random & 0.880 (0.866) / 28.8 (28.0) & 0.876 (0.862) / 28.4 (27.6) & 0.891 (0.877) / 29.6 (28.7) & 0.887 (0.873) / 29.1 (28.2) \\
        & Periodic & 0.877 (0.863) / 28.5 (27.7) & 0.873 (0.859) / 28.1 (27.3) & 0.888 (0.873) / 29.2 (28.4) & 0.884 (0.870) / 28.7 (27.9) \\
        \multirow{2}{*}{ICT} & Random & 0.872 (0.858) / 28.5 (27.6) & 0.868 (0.854) / 28.0 (27.2) & 0.882 (0.869) / 29.0 (28.2) & 0.876 (0.864) / 28.6 (27.7) \\
        & Periodic & 0.869 (0.855) / 28.1 (27.3) & 0.865 (0.851) / 27.7 (26.9) & 0.880 (0.866) / 28.7 (27.9) & 0.873 (0.862) / 28.3 (27.4) \\
        \multirow{2}{*}{Spline} & Random & 0.718 (0.704) / 24.7 (23.8) & 0.713 (0.699) / 24.3 (23.4) & 0.728 (0.714) / 25.3 (24.4) & 0.723 (0.710) / 24.9 (24.2) \\
        & Periodic & 0.715 (0.701) / 24.4 (23.5) & 0.710 (0.696) / 24.0 (23.1) & 0.726 (0.711) / 25.0 (24.1) & 0.721 (0.707) / 24.5 (23.7) \\
        \multirow{2}{*}{Bilinear} & Random & 0.711 (0.697) / 23.4 (22.6) & 0.707 (0.693) / 23.0 (22.2) & 0.722 (0.708) / 24.1 (23.2) & 0.718 (0.702) / 23.7 (22.7) \\
        & Periodic & 0.708 (0.694) / 23.1 (22.3) & 0.704 (0.690) / 22.7 (21.9) & 0.718 (0.704) / 23.7 (22.8) & 0.714 (0.700) / 23.3 (22.4) \\
        \bottomrule
    \end{tabular}}
    \caption{SSIM / PSNR on {\tt TomoBank} and {\tt LoDoPaB} datasets under random and periodic masks (ratio = 0.6, 0.8) at resolutions 2048$\times$2048 and 1024$\times$1024. Each entry is formatted as: completed sinogram (CT reconstruction by FBP).}
    \label{tab:accuracy}
\end{table*}

\noindent{\bf Evaluation Platform and Settings.}
All experiments are performed on a single NVIDIA A100 GPU with 40 GB on-device memory from the Polaris supercomputer of Argonne Leadership Computing Facility (ALCF) resources at Argonne National Laboratory (ANL), using CUDA 12.2, and PyTorch 2.1.0. All implementations employ PyTorch built-in optimizations: \texttt{torch.compile()} to enable graph-mode execution, mixed-precision inference via \texttt{torch.autocast()} (fp16), and \texttt{cudnn.benchmark = True} for kernel tuning.

All diffusion-based experiments use the denoising diffusion implicit models (DDIM) \cite{song2020denoising} sampler with 50 steps. This step count is consistent with values evaluated in DDIM paper, where 50-step sampling achieves a strong balance between quality and efficiency. All inferences are performed with a batch size of 1. For high-resolution inference, we partition sinograms into overlapping patches with patch size $P=128$, overlap $O=32$, and stride $S=96$. This setting balances efficiency and contextual coverage, while overlap mitigates boundary artifacts. For patch skipping, the high-frequency threshold is set to $\tau = 0.08$, which empirically filters low-information patches without affecting completion fidelity. For adaptive step scheduling, we use $(S_{min},S_{max}) = (10,50)$ to stay within the DDIM step budget while allowing fine-grained inference depth. More details about hyperparameters are in the extended version.

To ensure fair comparison, all baseline models are retrained from scratch using their official codebases. Since~\modelname~operates entirely at inference time, no retraining or task-specific fine-tuning is applied. Each dataset provides 100k samples, with 80\% used for training at a resolution of 512$\times$512, 10\% for validation, and 10\% for testing.

\noindent{\bf Dataset.}
We evaluate two real-world datasets {\tt TomoBank} and {\tt LoDoPaB}. {\tt TomoBank} \cite{de2018tomobank} consists of real sinogram images obtained from various materials and objects, derived from actual synchrotron radiation CT experiments, including Advanced Photon Source (APS) at ANL. It also includes samples with dynamic features \cite{mohan2015timbir} and in situ \cite{pelt2013fast} measurements, which capture a wide range of experiments at APS and other synchrotron facilities. To ensure consistency, we follow the official TomoPy \cite{gursoy2014tomopy} preprocessing pipeline, which includes ring artifact removal, rotation center alignment, and normalization of intensities to the [0,1] range. Projection counts, angular sampling, and rotation centers are configured according to the metadata provided in TomoBank. {\tt LoDoPaB} \cite{leuschner2021lodopab} is derived from the publicly available LIDC-IDRI lung CT database \cite{armato2011lung}. It provides simulated sinograms generated with a fixed parallel-beam geometry (1,000 projection angles, 513 detector bins), paired with real patient CT images. Unlike {\tt TomoBank}, where sinograms are physically measured, {\tt LoDoPaB} contains numerically generated sinograms with official splits provided via the DIVAL benchmark library \cite{DIVAL}. To simulate realistic sparse-view CT scenarios, we apply two types of angular masks: (1) random masks, which discard a random subset of projection angles, and (2) periodic masks, which regularly subsample angles at fixed intervals. The mask ratio denotes the fraction of angles removed (e.g., 0.8 means only 20\% of the projection angles are retained). In this work, we mainly evaluate challenging cases with mask ratios of 0.6 and 0.8, where the majority of angular information is missing.

\begin{figure*}[t]
    \centering
    
    \begin{subfigure}{0.13\textwidth}
        \centering
        \includegraphics[width=\textwidth]{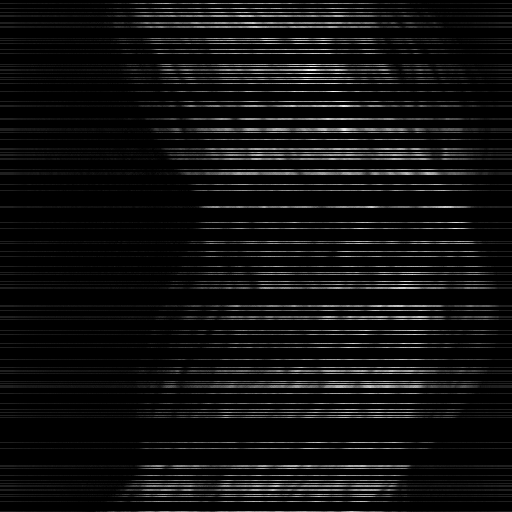}
    \end{subfigure}
    \begin{subfigure}{0.13\textwidth}
        \centering
        \includegraphics[width=\textwidth]{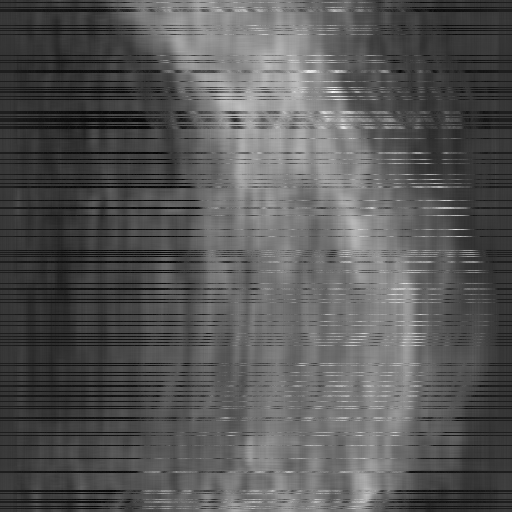}
    \end{subfigure}
    \begin{subfigure}{0.13\textwidth}
        \centering
        \includegraphics[width=\textwidth]{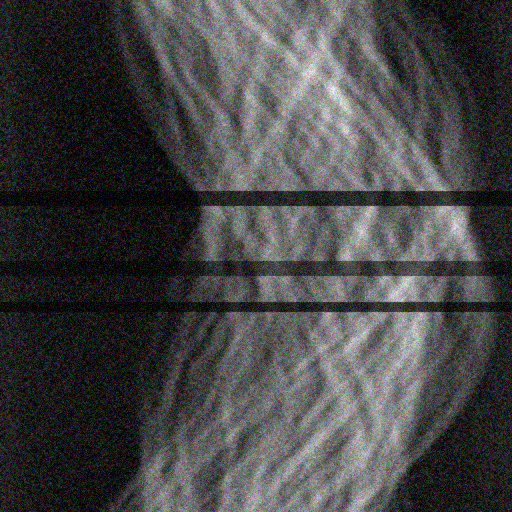}
    \end{subfigure}
    \begin{subfigure}{0.13\textwidth}
        \centering
        \includegraphics[width=\textwidth]{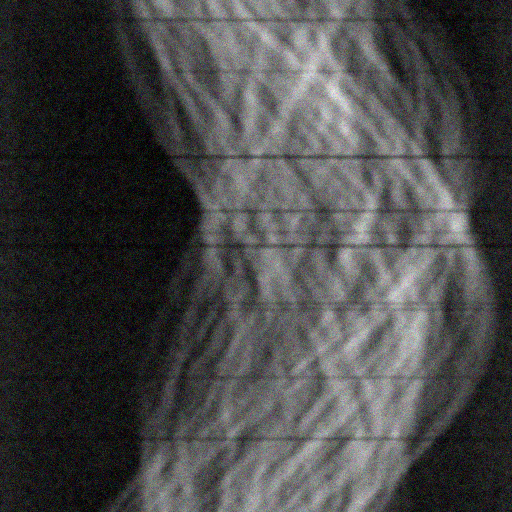}
    \end{subfigure}
    \begin{subfigure}{0.13\textwidth}
        \centering
        \includegraphics[width=\textwidth]{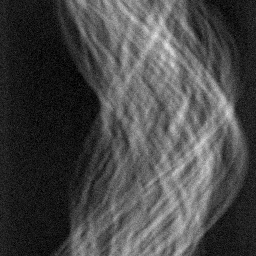}
    \end{subfigure}
    \begin{subfigure}{0.13\textwidth}
        \centering
        \includegraphics[width=\textwidth]{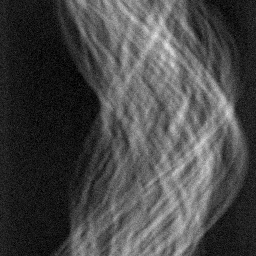}
    \end{subfigure}
    \begin{subfigure}{0.13\textwidth}
        \centering
        \includegraphics[width=\textwidth]{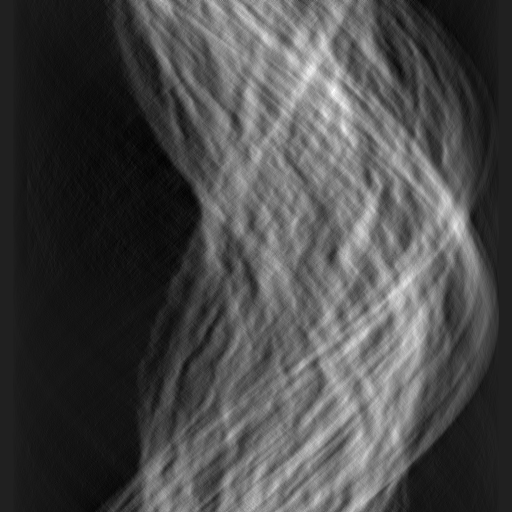}
    \end{subfigure}
    
    \begin{subfigure}{0.13\textwidth}
        \centering
        \includegraphics[width=\textwidth]{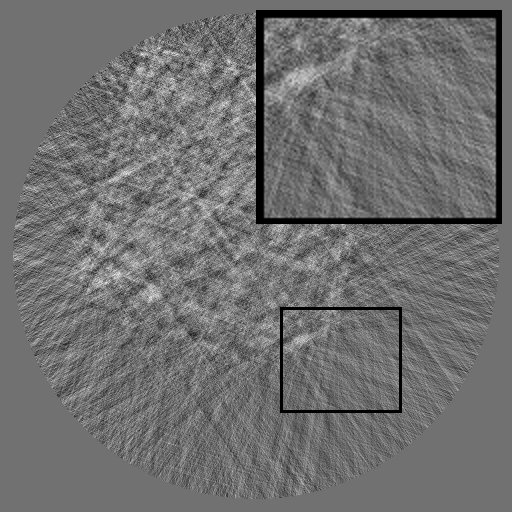}
    \end{subfigure}
    \begin{subfigure}{0.13\textwidth}
        \centering
        \includegraphics[width=\textwidth]{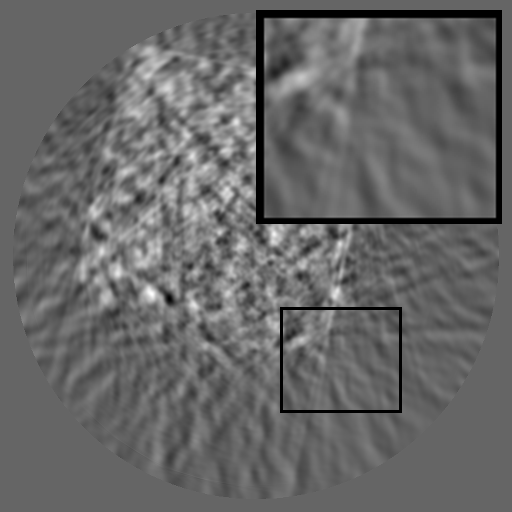}
    \end{subfigure}
    \begin{subfigure}{0.13\textwidth}
        \centering
        \includegraphics[width=\textwidth]{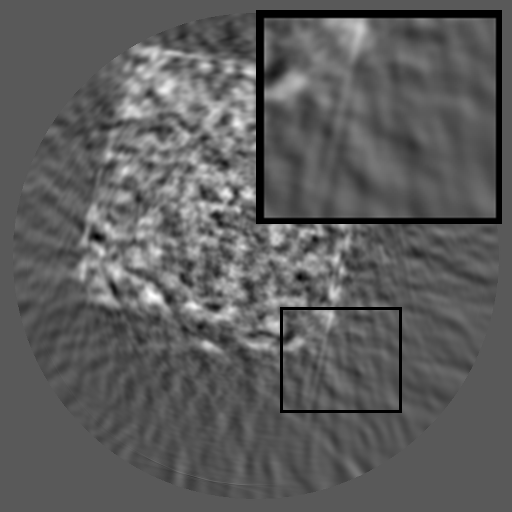}
    \end{subfigure}
    \begin{subfigure}{0.13\textwidth}
        \centering
        \includegraphics[width=\textwidth]{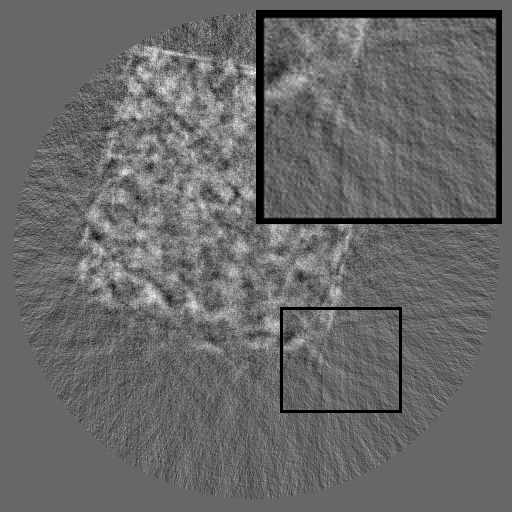}
    \end{subfigure}
    \begin{subfigure}{0.13\textwidth}
        \centering
        \includegraphics[width=\textwidth]{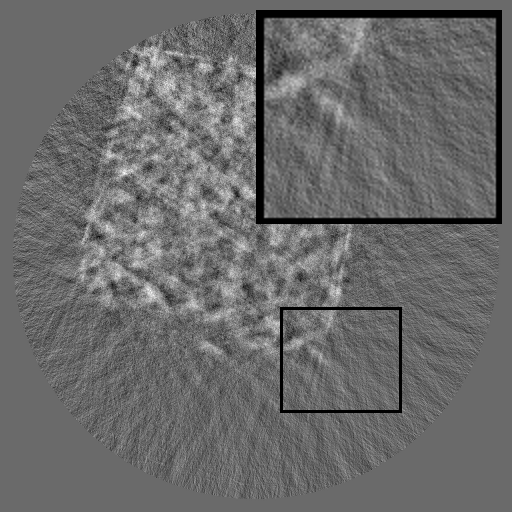}
    \end{subfigure}
    \begin{subfigure}{0.13\textwidth}
        \centering
        \includegraphics[width=\textwidth]{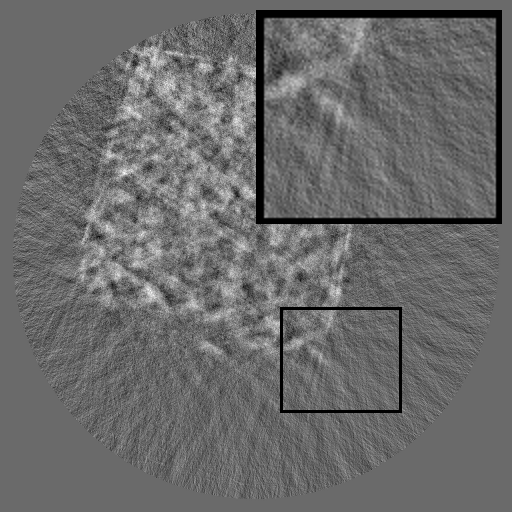}
    \end{subfigure}
    \begin{subfigure}{0.13\textwidth}
        \centering
        \includegraphics[width=\textwidth]{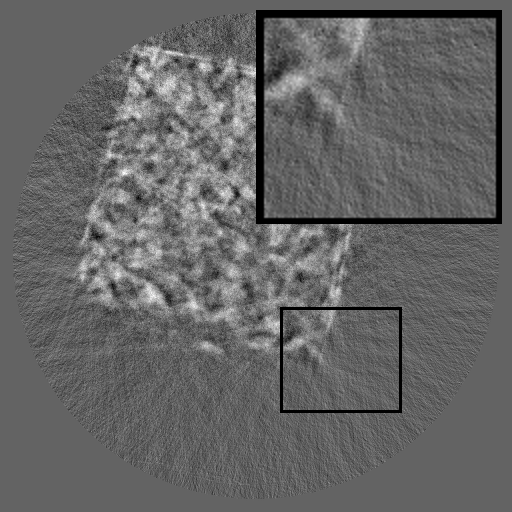}
    \end{subfigure}

    \begin{subfigure}{0.13\textwidth}
        \centering
        \includegraphics[width=\textwidth]{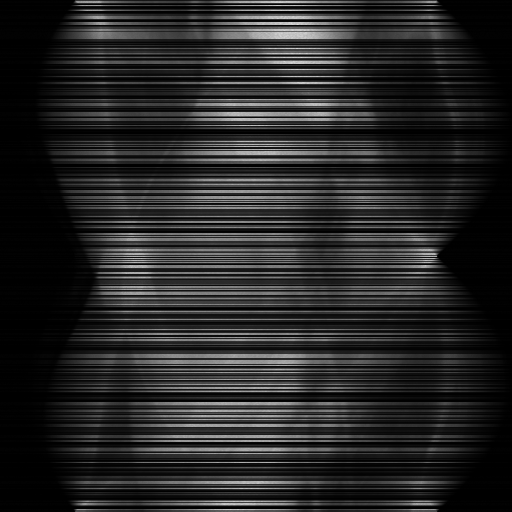}
    \end{subfigure}
    \begin{subfigure}{0.13\textwidth}
        \centering
        \includegraphics[width=\textwidth]{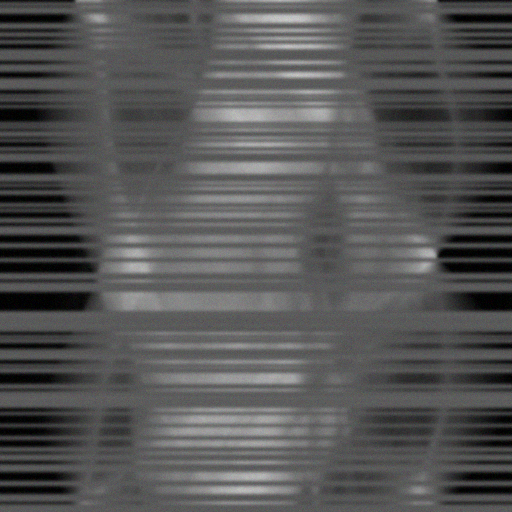}
    \end{subfigure}
    \begin{subfigure}{0.13\textwidth}
        \centering
        \includegraphics[width=\textwidth]{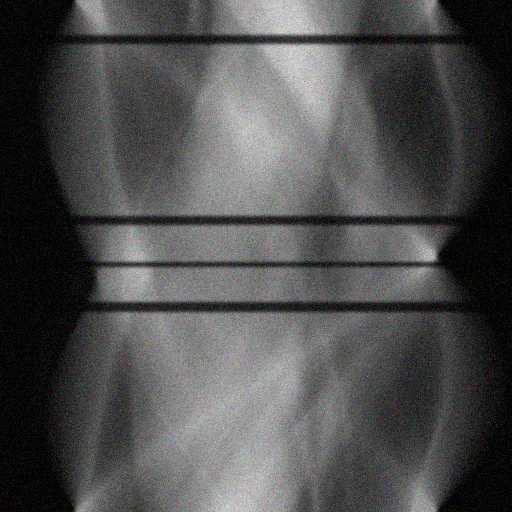}
    \end{subfigure}
    \begin{subfigure}{0.13\textwidth}
        \centering
        \includegraphics[width=\textwidth]{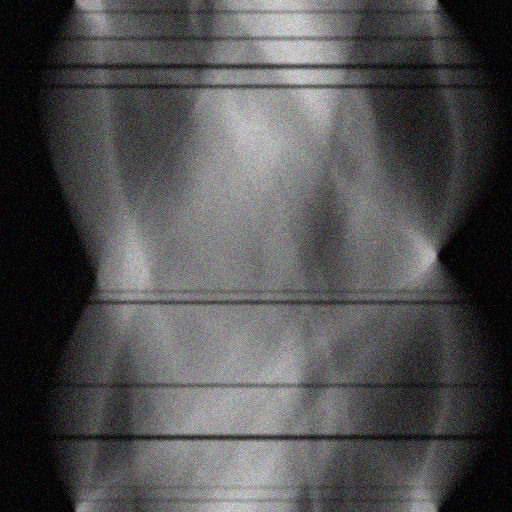}
    \end{subfigure}
    \begin{subfigure}{0.13\textwidth}
        \centering
        \includegraphics[width=\textwidth]{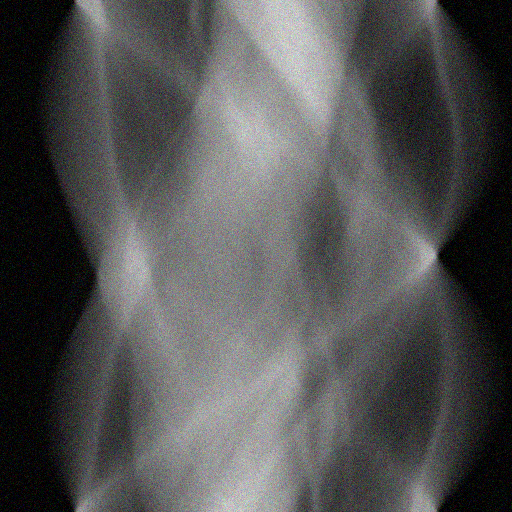}
    \end{subfigure}
    \begin{subfigure}{0.13\textwidth}
        \centering
        \includegraphics[width=\textwidth]{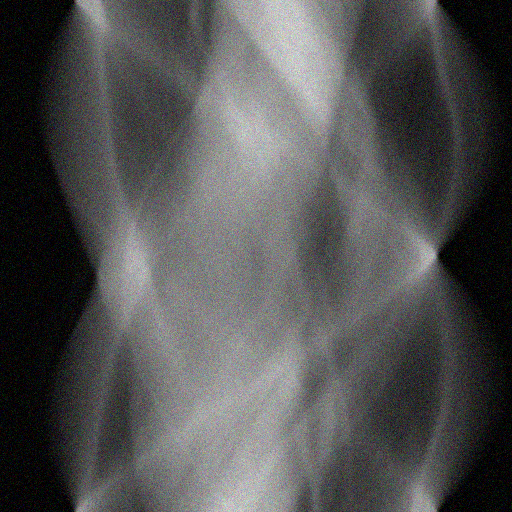}
    \end{subfigure}
    \begin{subfigure}{0.13\textwidth}
        \centering
        \includegraphics[width=\textwidth]{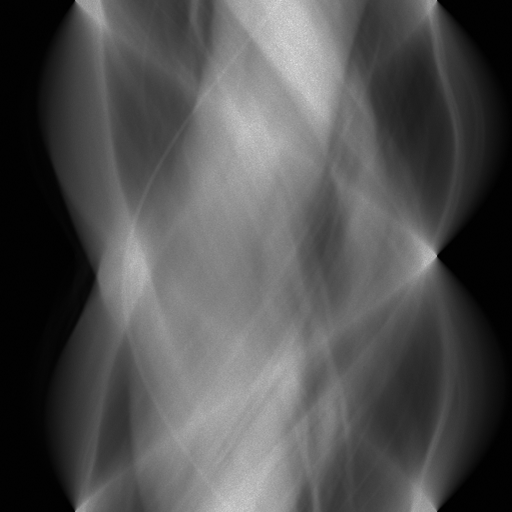}
    \end{subfigure}

    \begin{subfigure}{0.13\textwidth}
        \centering
        \includegraphics[width=\textwidth]{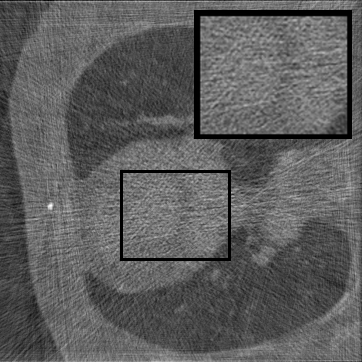}
        \captionsetup{labelformat=empty}
        \caption{Mask}
    \end{subfigure}
    \begin{subfigure}{0.13\textwidth}
        \centering
        \includegraphics[width=\textwidth]{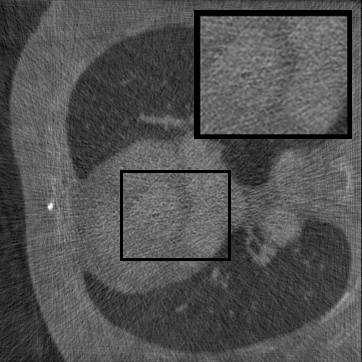}
        \captionsetup{labelformat=empty}
        \caption{SinoTx}
    \end{subfigure}
    \begin{subfigure}{0.13\textwidth}
        \centering
        \includegraphics[width=\textwidth]{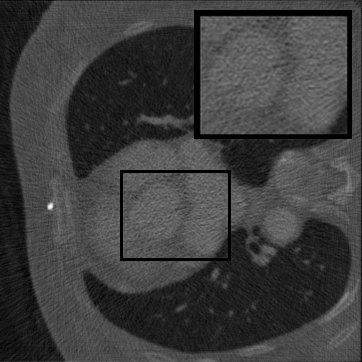}
        \captionsetup{labelformat=empty}
        \caption{DiffIR}
    \end{subfigure}
    \begin{subfigure}{0.13\textwidth}
        \centering
        \includegraphics[width=\textwidth]{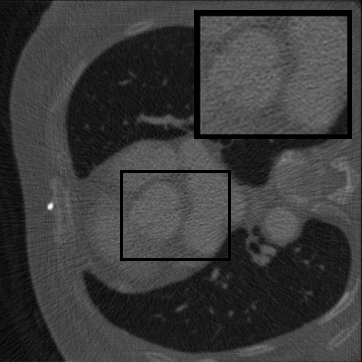}
        \captionsetup{labelformat=empty}
        \caption{HiDiffusion}
    \end{subfigure}
    \begin{subfigure}{0.13\textwidth}
        \centering
        \includegraphics[width=\textwidth]{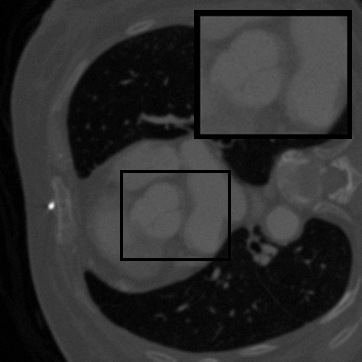}
        \captionsetup{labelformat=empty}
        \caption{RePaint}
    \end{subfigure}
    \begin{subfigure}{0.13\textwidth}
        \centering
        \includegraphics[width=\textwidth]{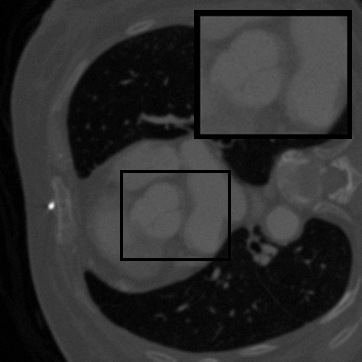}
        \captionsetup{labelformat=empty}
        \caption{HRSino}
    \end{subfigure}
    \begin{subfigure}{0.13\textwidth}
        \centering
        \includegraphics[width=\textwidth]{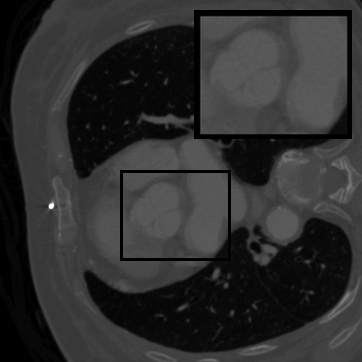}
        \captionsetup{labelformat=empty}
        \caption{Ground Truth}
    \end{subfigure}
    \caption{Qualitative completion results on \texttt{TomoBank} (lines 1 to 2) and \texttt{LoDoPaB} (lines 3 to 4) with random mask (ratio = 0.8) at 1024$\times$1024 resolution. Odd columns and even columns show the sinograms and reconstructed images, respectively.}
    \label{fig:qualitative}
\end{figure*}

\noindent{\bf Metrics.}
We evaluate each method using both efficiency and quality metrics. For computational performance, we report peak GPU memory usage and inference runtime, measured over multiple forward passes. Peak memory reflects the maximum GPU allocation during inference, indicating worst-case hardware demand. For completion fidelity, we use Structural Similarity Index (SSIM)~\cite{wang2004image} and Peak Signal-to-Noise Ratio (PSNR). SSIM captures perceptual and structural consistency, while PSNR quantifies absolute pixel-wise differences. All images are normalized to the [0, 1] range before metric computation. SSIM is computed in single-scale mode using a 11$\times$11 Gaussian kernel ($\sigma =1.5$). All metrics are computed on grayscale images. 

\noindent{\bf Baselines.}
We evaluate~\modelname~against representative baselines, including diffusion-based models optimized for high-resolution completion, transformer-based methods, and classical techniques. Among diffusion models, we consider RePaint \cite{lugmayr2022repaint}, a general-purpose completion model that serves as the backbone for~\modelname; HiDiffusion \cite{zhang2024hidiffusion}, which improves memory efficiency through resolution-aware U-Net design and attention; DiffIR \cite{xia2023diffir}, which introduces a intermediate prior to reduce sampling steps for image restoration; MCG \cite{chung2022improving}, which improves diffusion sampling stability via gradient-based manifold consistency guidance; and TD-Paint \cite{DBLP:conf/iclr/MayetS0GH025}, which accelerates diffusion-based completion by reusing known regions across diffusion timesteps without architectural modification. For transformer-based approaches, we include SinoTx \cite{jiaze2025sinotx}, tailored to sinogram completion, and ICT \cite{wan2021high}, designed for efficient completion. Finally, Bilinear and Spline interpolation serve as classical non-learning baselines.

\subsection{Overall Quantitative and Qualitative Results}
\label{sec:overall}

\subsubsection{Peak Memory Usage Comparison }

As shown in Tab~\ref{tab:memory-runtime},~\modelname~significantly reduces peak GPU memory usage compared to prior methods across both tested resolutions. RePaint, DiffIR, MCG, and TD-Paint all encounter out-of-memory (OOM) failures at 2048$\times$2048 resolution, indicating that standard diffusion pipelines cannot scale to such resolutions even with a high-end GPU. In contrast,~\modelname~remains fully operational under these conditions and achieves up to a 30.81\% reduction in memory usage relative to the most efficient baseline HiDiffusion. Notably, this trend holds consistently across input sizes, indicating that our framework scales gracefully.

\subsubsection{Inference Time Comparison}

As shown in Tab~\ref{tab:memory-runtime},~\modelname~consistently achieves faster inference than the fastest diffusion-based baseline, HiDiffusion, with speed up to 17.58\% at 2048$\times$2048 resolution. These improvements stem from two sources: progressive scheduling reduces redundant steps in simple regions, while patch skipping eliminates computation entirely for spectrally sparse areas. We further validate the effectiveness of these two mechanisms in the ablation studies (Sec~\ref{sec:ablation}). This demonstrates that high-resolution completion can be accelerated without compromising completion quality.

\subsubsection{Completion Quality}

Tab~\ref{tab:accuracy} summarizes accuracy results. At 2048$\times$2048 resolution,~\modelname~achieves the best performance among all baselines while remaining memory-efficient, demonstrating its ability to extend high-quality completion to resolutions where other diffusion models fail. At 1024$\times$1024,~\modelname~delivers accuracy comparable to its computation-intensive counterpart RePaint, showing that our optimizations do not compromise fidelity at moderate scales. Compared to DiffIR, MCG, TD-Paint, and HiDiffusion,~\modelname~consistently achieves higher SSIM and PSNR across mask ratios, with improvements up to +0.03 SSIM and +1.8 dB PSNR. Fig~\ref{fig:qualitative} visualizes sinogram completion and reconstructed images, where~\modelname~produces nearly indistinguishable results from RePaint. These findings confirm that~\modelname~fundamentally extends diffusion-based completion to 2048$\times$2048 resolution in a more memory- and runtime-efficient manner.

\begin{table}
    \centering
    {\scriptsize
    \begin{tabular}{lcccc}
        \toprule
        Method & Mem & Inf time & SSIM & PSNR \\
        \midrule
        \modelname & 24.7 & 2.25 & 0.927 & 30.6 \\
        W/o low & 24.7 & 2.15 & 0.901 & 28.7 \\
        W/o mid & 15.0 & 1.80 & 0.893 & 28.2 \\
        W/o high & 24.5 & 1.20 & 0.870 & 27.1 \\
        W/o adaptive steps & 24.7 & 3.05 & 0.928 & 30.9 \\
        W/o patch skipping & 24.7 & 3.25 & 0.929 & 30.9 \\
        \bottomrule
    \end{tabular}}
    \caption{Ablation study of multi-resolutions, adaptive denoising, and patch skipping. Results are reported as peak GPU memory (GB), inference time (s), and SSIM/PSNR on {\tt TomoBank} with random masks (ratio = 0.8) at resolution 2048$\times$2048.}
    \label{tab:ablation}
\end{table}

\subsection{Ablation Studies}
\label{sec:ablation}

To analyze the impact of individual components in~\modelname, we conduct ablation studies on its three core mechanisms: multi-resolution progressive inference, adaptive denoising based on structural complexity, and sparse patch skipping based on spectral sparsity. Tab~\ref{tab:ablation} summarizes the results. Removing any resolution stage substantially degrades completion quality, indicating that different resolutions provide complementary benefits and that all stages are necessary to maintain overall fidelity. Peak memory is dominated by the mid-resolution stage, since high-resolution inference is performed patch-wise and patches are processed sequentially. Removing adaptive denoising or patch skipping does not affect SSIM/PSNR, but results in significant efficiency loss: inference time increases by 35.6\% without adaptive steps (all set to 50) and by 44.4\% without patch skipping. Overall, these results demonstrate the complementary roles of three mechanisms in enabling efficient high-resolution completion.

\subsection{Compatibility with existing optimizations}

\begin{table}[t]
    \centering
    {\scriptsize
    \begin{tabular}{lcccc}
        \toprule
        Method & Mem & Inf time & SSIM & PSNR \\
        \midrule
        HiDiffusion & 35.7 & 2.73 & 0.900 & 28.8 \\
        HiDiffusion+\modelname & 20.0 & 1.85 & 0.899 & 28.8 \\
        DiffIR & \multicolumn{4}{c}{OOM} \\
        DiffIR+\modelname & 24.5 & 2.0 & 0.883 & 28.6 \\
        \bottomrule
    \end{tabular}}
    \caption{Orthogonality study on {\tt TomoBank} with random masks (ratio = 0.8) at resolution 2048$\times$2048.}
    \label{tab:orthogonality}
\end{table}

Our optimizations are orthogonal to existing efficiency-oriented diffusion designs.~\modelname~is built upon RePaint, but our optimizations can be applied to other diffusion architectures. We combine~\modelname~with two recent baselines, HiDiffusion and DiffIR. As shown in Tab~\ref{tab:orthogonality}, integrating~\modelname~further reduces peak memory and inference time, while maintaining comparable completion quality.

\section{Conclusion}

We present~\modelname, a novel efficient high-resolution sinogram completion framework.~\modelname~integrates mechanisms that adaptively allocate computation across resolutions, spatial regions, and inference depth. This design enables completion on 2048$\times$2048 sinograms using a single GPU, significantly reducing memory usage and runtime while maintaining the completion fidelity.

\clearpage

\section*{Acknowledgments}

The authors would like to thank the anonymous reviewers for their constructive comments and suggestions. This work was supported in part by the National Science Foundation (NSF) under awards CCF-2047516 (CAREER), CCF-2146873, CNS-2230944, IIS-2451436, and OAC-2403088, National Institutes of Health (NIH) under the award of R01 AG085616, Commonwealth Cyber Initiative (CCI) under the award HC-2Q26-032, and the U.S. Department of Energy (DOE) under Contract No. DE-AC02-06CH11357, including funding from the XSCOPE project and Laboratory Directed Research and Development Program (Project No. 2023-0104) at Argonne National Laboratory (ANL). It used resources of the Argonne Leadership Computing Facility (ALCF) and the Advanced Photon Source (APS), both DOE Office of Science user facilities at ANL, and was based on research supported by the U.S. DOE Office of Science ASCR Program under Contract No. DE-AC02-06CH11357. Any opinions, findings, and conclusions or recommendations expressed in this material are those of the authors and do not necessarily reflect the views of the funding agencies.

\bibliographystyle{named}
\bibliography{ijcai26}

@book{kalender2011computed,
  title={Computed tomography: fundamentals, system technology, image quality, applications},
  author={Kalender, Willi A},
  year={2011},
  publisher={John Wiley \& Sons}
}

@inproceedings{sohl2015deep,
  title={Deep unsupervised learning using nonequilibrium thermodynamics},
  author={Sohl-Dickstein, Jascha and Weiss, Eric and Maheswaranathan, Niru and Ganguli, Surya},
  booktitle={International conference on machine learning},
  pages={2256--2265},
  year={2015},
  organization={pmlr}
}

@article{ho2020denoising,
  title={Denoising diffusion probabilistic models},
  author={Ho, Jonathan and Jain, Ajay and Abbeel, Pieter},
  journal={Advances in neural information processing systems},
  volume={33},
  pages={6840--6851},
  year={2020}
}

@inproceedings{lugmayr2022repaint,
  title={Repaint: Inpainting using denoising diffusion probabilistic models},
  author={Lugmayr, Andreas and Danelljan, Martin and Romero, Andres and Yu, Fisher and Timofte, Radu and Van Gool, Luc},
  booktitle={Proceedings of the IEEE/CVF conference on computer vision and pattern recognition},
  pages={11461--11471},
  year={2022}
}

@inproceedings{saharia2022palette,
  title={Palette: Image-to-image diffusion models},
  author={Saharia, Chitwan and Chan, William and Chang, Huiwen and Lee, Chris and Ho, Jonathan and Salimans, Tim and Fleet, David and Norouzi, Mohammad},
  booktitle={ACM SIGGRAPH 2022 conference proceedings},
  pages={1--10},
  year={2022}
}

@inproceedings{avrahami2022blended,
  title={Blended diffusion for text-driven editing of natural images},
  author={Avrahami, Omri and Lischinski, Dani and Fried, Ohad},
  booktitle={Proceedings of the IEEE/CVF conference on computer vision and pattern recognition},
  pages={18208--18218},
  year={2022}
}

@article{zhang2023towards,
  title={Towards coherent image inpainting using denoising diffusion implicit models},
  author={Zhang, Guanhua and Ji, Jiabao and Zhang, Yang and Yu, Mo and Jaakkola, Tommi S and Chang, Shiyu},
  year={2023}
}

@inproceedings{tumanyan2023plug,
  title={Plug-and-play diffusion features for text-driven image-to-image translation},
  author={Tumanyan, Narek and Geyer, Michal and Bagon, Shai and Dekel, Tali},
  booktitle={Proceedings of the IEEE/CVF Conference on Computer Vision and Pattern Recognition},
  pages={1921--1930},
  year={2023}
}

@inproceedings{zhang2023adding,
  title={Adding conditional control to text-to-image diffusion models},
  author={Zhang, Lvmin and Rao, Anyi and Agrawala, Maneesh},
  booktitle={Proceedings of the IEEE/CVF international conference on computer vision},
  pages={3836--3847},
  year={2023}
}

@article{shannon1948mathematical,
  title={A mathematical theory of communication},
  author={Shannon, Claude E},
  journal={The Bell system technical journal},
  volume={27},
  number={3},
  pages={379--423},
  year={1948},
  publisher={Nokia Bell Labs}
}

@book{slaney1988principles,
  title={Principles of computerized tomographic imaging},
  author={Slaney, Malcolm and Kak, AC},
  year={1988},
  publisher={IEEE press}
}

@inproceedings{zhang2024hidiffusion,
  title={Hidiffusion: Unlocking higher-resolution creativity and efficiency in pretrained diffusion models},
  author={Zhang, Shen and Chen, Zhaowei and Zhao, Zhenyu and Chen, Yuhao and Tang, Yao and Liang, Jiajun},
  booktitle={European Conference on Computer Vision},
  pages={145--161},
  year={2024},
  organization={Springer}
}

@inproceedings{xia2023diffir,
  title={Diffir: Efficient diffusion model for image restoration},
  author={Xia, Bin and Zhang, Yulun and Wang, Shiyin and Wang, Yitong and Wu, Xinglong and Tian, Yapeng and Yang, Wenming and Van Gool, Luc},
  booktitle={Proceedings of the IEEE/CVF International Conference on Computer Vision},
  pages={13095--13105},
  year={2023}
}

@article{de2018tomobank,
  title={TomoBank: a tomographic data repository for computational x-ray science},
  author={De Carlo, Francesco and G{\"u}rsoy, Do{\u{g}}a and Ching, Daniel J and Batenburg, K Joost and Ludwig, Wolfgang and Mancini, Lucia and Marone, Federica and Mokso, Rajmund and Pelt, Dani{\"e}l M and Sijbers, Jan and others},
  journal={Measurement Science and Technology},
  volume={29},
  number={3},
  pages={034004},
  year={2018},
  publisher={IOP Publishing}
}

@article{pelt2013fast,
  title={Fast tomographic reconstruction from limited data using artificial neural networks},
  author={Pelt, Daniel Maria and Batenburg, Kees Joost},
  journal={IEEE Transactions on Image Processing},
  volume={22},
  number={12},
  pages={5238--5251},
  year={2013},
  publisher={IEEE}
}

@article{mohan2015timbir,
  title={TIMBIR: A method for time-space reconstruction from interlaced views},
  author={Mohan, K. Aditya and Venkatakrishnan, SV and Gibbs, John W and Gulsoy, Emine Begum and Xiao, Xianghui and De Graef, Marc and Voorhees, Peter W and Bouman, Charles A},
  journal={IEEE Transactions on Computational Imaging},
  volume={1},
  number={2},
  pages={96--111},
  year={2015},
  publisher={IEEE}
}

@article{gursoy2014tomopy,
  title={TomoPy: a framework for the analysis of synchrotron tomographic data},
  author={G{\"u}rsoy, Doga and De Carlo, Francesco and Xiao, Xianghui and Jacobsen, Chris},
  journal={Journal of synchrotron radiation},
  volume={21},
  number={5},
  pages={1188--1193},
  year={2014},
  publisher={International Union of Crystallography}
}

@article{wang2004image,
  title={Image quality assessment: from error visibility to structural similarity},
  author={Wang, Zhou and Bovik, Alan C and Sheikh, Hamid R and Simoncelli, Eero P},
  journal={IEEE transactions on image processing},
  volume={13},
  number={4},
  pages={600--612},
  year={2004},
  publisher={IEEE}
}

@article{lee2018deep,
  title={Deep-neural-network-based sinogram synthesis for sparse-view CT image reconstruction},
  author={Lee, Hoyeon and Lee, Jongha and Kim, Hyeongseok and Cho, Byungchul and Cho, Seungryong},
  journal={IEEE Transactions on Radiation and Plasma Medical Sciences},
  volume={3},
  number={2},
  pages={109--119},
  year={2018},
  publisher={IEEE}
}

@article{jin2017deep,
  title={Deep convolutional neural network for inverse problems in imaging},
  author={Jin, Kyong Hwan and McCann, Michael T and Froustey, Emmanuel and Unser, Michael},
  journal={IEEE transactions on image processing},
  volume={26},
  number={9},
  pages={4509--4522},
  year={2017},
  publisher={IEEE}
}

@article{salimans2022progressive,
  title={Progressive distillation for fast sampling of diffusion models},
  author={Salimans, Tim and Ho, Jonathan},
  journal={arXiv preprint arXiv:2202.00512},
  year={2022}
}

@article{xiao2024fastcomposer,
  title={Fastcomposer: Tuning-free multi-subject image generation with localized attention},
  author={Xiao, Guangxuan and Yin, Tianwei and Freeman, William T and Durand, Fr{\'e}do and Han, Song},
  journal={International Journal of Computer Vision},
  pages={1--20},
  year={2024},
  publisher={Springer}
}

@article{avrahami2023blended,
  title={Blended latent diffusion},
  author={Avrahami, Omri and Fried, Ohad and Lischinski, Dani},
  journal={ACM transactions on graphics (TOG)},
  volume={42},
  number={4},
  pages={1--11},
  year={2023},
  publisher={ACM New York, NY, USA}
}

@article{chen2016training,
  title={Training deep nets with sublinear memory cost},
  author={Chen, Tianqi and Xu, Bing and Zhang, Chiyuan and Guestrin, Carlos},
  journal={arXiv preprint arXiv:1604.06174},
  year={2016}
}

@article{jain2020checkmate,
  title={Checkmate: Breaking the memory wall with optimal tensor rematerialization},
  author={Jain, Paras and Jain, Ajay and Nrusimha, Aniruddha and Gholami, Amir and Abbeel, Pieter and Gonzalez, Joseph and Keutzer, Kurt and Stoica, Ion},
  journal={Proceedings of Machine Learning and Systems},
  volume={2},
  pages={497--511},
  year={2020}
}

@article{xia2021fully,
  title={Fully dynamic inference with deep neural networks},
  author={Xia, Wenhan and Yin, Hongxu and Dai, Xiaoliang and Jha, Niraj K},
  journal={IEEE Transactions on Emerging Topics in Computing},
  volume={10},
  number={2},
  pages={962--972},
  year={2021},
  publisher={IEEE}
}

@article{yuan2021mest,
  title={Mest: Accurate and fast memory-economic sparse training framework on the edge},
  author={Yuan, Geng and Ma, Xiaolong and Niu, Wei and Li, Zhengang and Kong, Zhenglun and Liu, Ning and Gong, Yifan and Zhan, Zheng and He, Chaoyang and Jin, Qing and others},
  journal={Advances in Neural Information Processing Systems},
  volume={34},
  pages={20838--20850},
  year={2021}
}

@inproceedings{meng2023distillation,
  title={On distillation of guided diffusion models},
  author={Meng, Chenlin and Rombach, Robin and Gao, Ruiqi and Kingma, Diederik and Ermon, Stefano and Ho, Jonathan and Salimans, Tim},
  booktitle={Proceedings of the IEEE/CVF Conference on Computer Vision and Pattern Recognition},
  pages={14297--14306},
  year={2023}
}

@article{li2023snapfusion,
  title={Snapfusion: Text-to-image diffusion model on mobile devices within two seconds},
  author={Li, Yanyu and Wang, Huan and Jin, Qing and Hu, Ju and Chemerys, Pavlo and Fu, Yun and Wang, Yanzhi and Tulyakov, Sergey and Ren, Jian},
  journal={Advances in Neural Information Processing Systems},
  volume={36},
  pages={20662--20678},
  year={2023}
}

@article{zhang2024effortless,
  title={Effortless Efficiency: Low-Cost Pruning of Diffusion Models},
  author={Zhang, Yang and Jin, Er and Dong, Yanfei and Khakzar, Ashkan and Torr, Philip and Stegmaier, Johannes and Kawaguchi, Kenji},
  journal={arXiv preprint arXiv:2412.02852},
  year={2024}
}

@article{lu2022dpm,
  title={Dpm-solver: A fast ode solver for diffusion probabilistic model sampling in around 10 steps},
  author={Lu, Cheng and Zhou, Yuhao and Bao, Fan and Chen, Jianfei and Li, Chongxuan and Zhu, Jun},
  journal={Advances in Neural Information Processing Systems},
  volume={35},
  pages={5775--5787},
  year={2022}
}

@article{zhu2024dip,
  title={Dip-go: A diffusion pruner via few-step gradient optimization},
  author={Zhu, Haowei and Tang, Dehua and Liu, Ji and Lu, Mingjie and Zheng, Jintu and Peng, Jinzhang and Li, Dong and Wang, Yu and Jiang, Fan and Tian, Lu and others},
  journal={Advances in Neural Information Processing Systems},
  volume={37},
  pages={92581--92604},
  year={2024}
}

@article{song2020denoising,
  title={Denoising diffusion implicit models},
  author={Song, Jiaming and Meng, Chenlin and Ermon, Stefano},
  journal={arXiv preprint arXiv:2010.02502},
  year={2020}
}

@article{leuschner2021lodopab,
  title={LoDoPaB-CT, a benchmark dataset for low-dose computed tomography reconstruction},
  author={Leuschner, Johannes and Schmidt, Maximilian and Baguer, Daniel Otero and Maass, Peter},
  journal={Scientific Data},
  volume={8},
  number={1},
  pages={109},
  year={2021},
  publisher={Nature Publishing Group UK London}
}

@misc{DIVAL,
  title={Deep Inversion Validation Library},
  author={Johannes Leuschner and others},
  howpublished={\url{https://github.com/jleuschn/dival}},
  year={2025},
}

@article{jiaze2025sinotx,
  title={SinoTx: A Transformer-based Model for Sinogram Inpainting},
  author={E, Jiaze and Liu, Zhengchun and Bicer, Tekin and Banerjee, Srutarshi and Kettimuthu, Rajkumar and Ren, Bin and Foster, Ian T},
  journal={Electronic Imaging},
  volume={37},
  pages={1--6},
  year={2025},
  publisher={Society for Imaging Science and Technology}
}

@inproceedings{wan2021high,
  title={High-fidelity pluralistic image completion with transformers},
  author={Wan, Ziyu and Zhang, Jingbo and Chen, Dongdong and Liao, Jing},
  booktitle={Proceedings of the IEEE/CVF international conference on computer vision},
  pages={4692--4701},
  year={2021}
}

@article{armato2011lung,
  title={The lung image database consortium (LIDC) and image database resource initiative (IDRI): a completed reference database of lung nodules on CT scans},
  author={Armato III, Samuel G and McLennan, Geoffrey and Bidaut, Luc and McNitt-Gray, Michael F and Meyer, Charles R and Reeves, Anthony P and Zhao, Binsheng and Aberle, Denise R and Henschke, Claudia I and Hoffman, Eric A and others},
  journal={Medical physics},
  volume={38},
  number={2},
  pages={915--931},
  year={2011},
  publisher={Wiley Online Library}
}

@article{guo2025advancing,
  title={Advancing Limited-Angle CT Reconstruction Through Diffusion-Based Sinogram Completion},
  author={Guo, Jiaqi and Lopez-Tapia, Santiago and Katsaggelos, Aggelos K},
  journal={arXiv preprint arXiv:2505.19385},
  year={2025}
}

@article{chung2022improving,
  title={Improving diffusion models for inverse problems using manifold constraints},
  author={Chung, Hyungjin and Sim, Byeongsu and Ryu, Dohoon and Ye, Jong Chul},
  journal={Advances in Neural Information Processing Systems},
  volume={35},
  pages={25683--25696},
  year={2022}
}

@inproceedings{DBLP:conf/iclr/MayetS0GH025,
  author       = {Tsiry Mayet and
                  Pourya Shamsolmoali and
                  Simon Bernard and
                  Eric Granger and
                  Romain H{\'{e}}rault and
                  Cl{\'{e}}ment Chatelain},
  title        = {TD-Paint: Faster Diffusion Inpainting Through Time-Aware Pixel Conditioning},
  booktitle    = {The Thirteenth International Conference on Learning Representations,
                  {ICLR} 2025, Singapore, April 24-28, 2025},
  publisher    = {OpenReview.net},
  year         = {2025},
  url          = {https://openreview.net/forum?id=erWwBoR59l},
  bibsource    = {dblp computer science bibliography, https://dblp.org}
}

\clearpage
\appendix

\section{Additional Design Notes}

\subsection{Sensitivity of the Patch-Skipping Threshold}

\subsubsection{Rationale Behind the Patch-Skipping Threshold Selection}

The parameter $\tau$ controls the patch-skipping threshold in~\modelname. Specifically, it determines the spectral energy level below which a patch is considered uninformative and thus can be skipped during inference. A larger value of $\tau$ results in more aggressive skipping, while a smaller value leads to fewer patches being skipped and then higher inference time.

We set $\tau = 0.08$ in all our experiments based on three practical considerations:
\begin{itemize}
    \item Empirical coverage of low-frequency components: In typical sinogram inputs, a threshold of 0.08 allows us to skip patches with minimal high-frequency activity while still retaining structurally important information.
    \item Interpretability and reproducibility: Compared to more aggressive thresholds, the 0.08 setting avoids dropping structurally ambiguous or borderline patches.
    \item Stability across datasets and resolutions: We observed that $\tau = 0.08$ yields stable performance across all datasets, and across multiple input resolutions and mask ratios. This suggests that the choice is not overly sensitive to dataset-, input resolution- or mask ratio-specific properties.
\end{itemize}
These empirical observations collectively support the choice of $\tau = 0.08$ as a robust and balanced default.

\subsubsection{Sensitivity Analysis Experiments}

\begin{table}[ht]
    \centering
    \caption{Sensitivity of~\modelname~to different values of $\tau$ on {\tt TomoBank} with random masks (ratio = 0.8) at resolution 2048$\times$2048. SSIM and PSNR are formatted as: completed sinogram (reconstructed CT image by FBP).}
    \label{tab:tau}
    \begin{tabular}{ccc}
        \toprule
        $\tau$ & SSIM & PSNR \\
        \midrule
        0.12 & 0.912 (0.908) & 29.7 (28.5) \\
        0.08 & 0.927 (0.913) & 30.6 (29.7) \\
        0.05 & 0.928 (0.913) & 30.7 (29.7) \\
        \bottomrule
    \end{tabular}
\end{table}

Tab~\ref{tab:tau} reports evaluation results for $\tau$ sensitivity. We compare SSIM and PSNR under three representative values of $\tau$: 0.12, 0.08, and 0.05. These thresholds respectively correspond to aggressive skipping, default setting, and conservative skipping.

We observed that decreasing $\tau$ beyond 0.08 yields diminishing returns in completion accuracy, while causing an increase in inference time due to the larger number of patches retained. Therefore, we set $\tau = 0.08$ as a practical balance point—preserving computational efficiency without sacrificing output quality.

\subsection{Distinction Between Patch Skipping and Adaptive Step Allocation}

While both patch skipping and adaptive step allocation rely on local complexity analysis, they serve distinct purposes and operate under different assumptions. Patch skipping is a coarse-grained gating mechanism that identifies structurally trivial regions where full denoising can be omitted entirely. It uses a simple frequency-domain criterion—the ratio of high-frequency energy—to detect spectrally sparse patches dominated by very smooth or empty content. This decision is binary and conservative: a skipped patch will never undergo any denoising, so the threshold must be set cautiously to avoid quality degradation.

In contrast, adaptive step allocation aims to adjust the computational effort spent on patches that are retained for inference. Instead of deciding whether to process a patch, it modulates how many DDIM steps to apply based on a richer complexity score. This score combines Shannon entropy over pixel intensities with the total frequency energy (via the L1 norm of the FFT), capturing both spatial irregularity and spectral richness. The resulting step count is assigned through a smooth sigmoid mapping centered around the mean complexity across all patches, ensuring stable and data-aware inference depth scheduling.

While both mechanisms involve complexity estimation, their criteria, granularity, and functional roles are fundamentally different.

\section{Limitations: Computational Cost Analysis}

\begin{table}[ht]
    \centering
    \caption{FLOPs (G) for different methods on \texttt{TomoBank} with random masks (ratio = 0.8) at resolutions 2048$\times$2048 and 1024$\times$1024. All FLOPs are estimated via code-level calculation based on resolution, patch size, and denoising steps. These calculations are performed using standard layer-wise formulas, without reliance on hardware profiling.}
    \label{tab:limitation}
    \begin{tabular}{lccc}
        \toprule
        Method & 2048$\times$2048 & 1024$\times$1024 \\
        \midrule
        RePaint & \multicolumn{1}{c}{\multirow{2}{*}{OOM}} & 34.7 \\
        DiffIR & & 22.5 \\
        HiDiffusion & 112.6 & 28.4 \\
        HRSino & 118.3 & 29.6 \\
        \bottomrule
    \end{tabular}
\end{table}

While~\modelname~improves memory usage and inference speed through hierarchical scheduling and patch-level optimization, it comes with a moderate increase in total FLOPs compared to other efficiency-oriented baselines. As shown in Tab~\ref{tab:limitation},~\modelname~incurs more FLOPs than HiDiffusion and DiffIR. This reflects the structural overhead of progressive inference and adaptive patch processing. While the FLOPs slightly increase compared to HiDiffusion and DiffIR, this trade-off is necessary to support higher resolution sinogram completion, while maintaining high memory efficiency and completion quality.

\end{document}